\pdfoutput=1
\documentclass[11pt]{article}

\usepackage[preprint]{acl}

\usepackage{times}
\usepackage{latexsym}
\usepackage{microtype} 
\usepackage{booktabs}
\usepackage{listings}

\usepackage[T1]{fontenc}
\usepackage[utf8]{inputenc}

\usepackage{microtype}

\usepackage{inconsolata}

\usepackage{graphicx}
\usepackage{amsmath}
\usepackage{cleveref}
\usepackage{amsfonts}
\usepackage{hyperref}

\newcommand{\Dataset}{PaperSearchQA}

\usepackage{placeins}       % for \FloatBarrier

\title{\Dataset: Learning to Search and Reason over Scientific Papers with RLVR}
\author{%
  \textbf{James Burgess}$^{1}$\thanks{Correspondence: \texttt{jmhb@stanford.edu}}
  \quad 
  \textbf{Jan N. Hansen}$^{1}$
  \quad 
  \textbf{Duo Peng}$^{2}$
  \quad 
  \textbf{Yuhui Zhang}$^{1}$
  \quad 
  \textbf{Alejandro Lozano}$^{1}$\\
  \textbf{Min Woo Sun}$^{1}$
  \quad 
  \textbf{Emma Lundberg}$^{1,2,3}$
  \quad  
  \textbf{Serena Yeung-Levy}$^{1,2}$\\
$^{1}$Stanford University, $^{2}$Chan Zuckerberg Biohub Network , $^{3}$KTH Royal Institute of Technology \\
  \\
  \faGlobe~\href{https://jmhb0.github.io/PaperSearchQA}{Project Page} \qquad
  \faDatabase~\href{https://huggingface.co/collections/jmhb/papersearchqa}{Datasets}
  \qquad \faGithub~\href{https://github.com/jmhb0/PaperSearchQA}{Code}
}



\begin{document}
\maketitle

\begin{abstract}
Search agents are language models (LMs) that reason and search knowledge bases (or the web) to answer questions; recent methods supervise only the final answer accuracy using reinforcement learning with verifiable rewards (RLVR). Most RLVR search agents tackle general-domain QA, which limits their relevance to technical AI systems in science, engineering, and medicine. In this work we propose training agents to search and reason over scientific papers -- this tests technical question-answering, it is directly relevant to real scientists, and the capabilities will be crucial to future AI Scientist systems. Concretely, we release a search corpus of 16 million biomedical paper abstracts and construct a challenging factoid QA dataset called \Dataset\ with 60k samples answerable from the corpus, along with benchmarks. We train search agents in this environment to outperform non-RL retrieval baselines; we also perform further quantitative analysis and observe interesting agent behaviors like planning, reasoning, and self-verification. Our corpus, datasets, and benchmarks are usable with the popular Search-R1 codebase for RLVR training and released on \href{https://huggingface.co/collections/jmhb/papersearchqa}{Hugging Face}. Finally, our data creation methods are scalable and easily extendable to other scientific domains. 
\end{abstract}

\section{Introduction}

\begin{figure}[t]
    \centering
    \includegraphics[width=0.49\textwidth,page=3, trim={0 235 1225 0},clip]{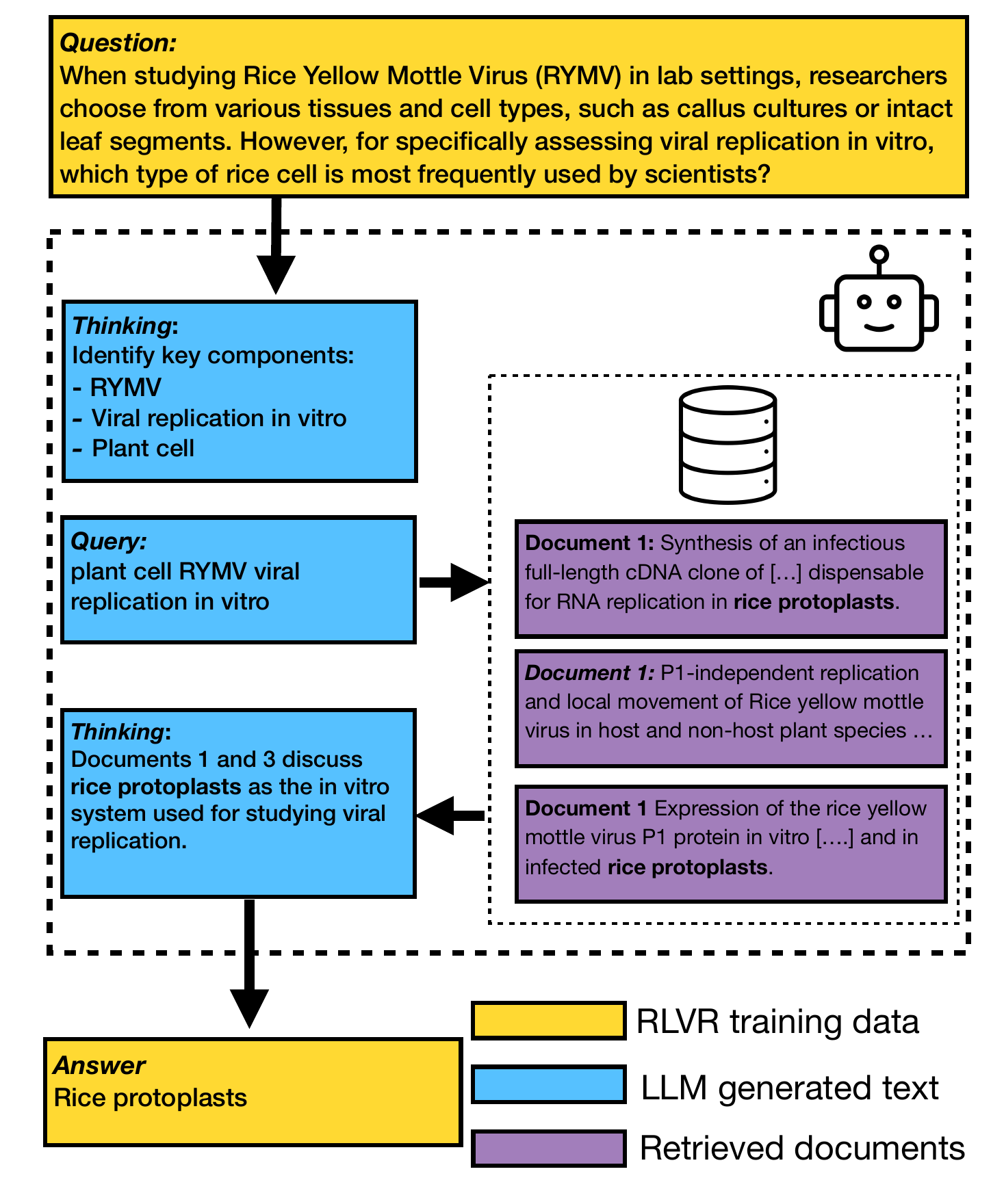}
    \caption{Search agents interleave reasoning and retrieval for question answering (QA). We study QA over scientific literature, contributing an environment for training agents with RL with verifiable rewards (RLVR). We release a training dataset of factoid QA (yellow boxes),  a retrieval corpus (purple), and benchmarks.
    }
    \label{fig:pull}
\end{figure}

Following the release of Deepseek-R1 \cite{guo2025deepseek} and OpenAI's o1 \cite{jaech2024openai}, much large language model (LLM) research has employed reinforcement learning with verifiable rewards (RLVR) \cite{shao2024deepseekmath, lambert2024t}. In RLVR, an LLM is prompted to answer a query, and a reward is given only if an automatic verifier deems the final output correct; the corresponding tokens are then used to update the model (\cref{app:rlvr} has more details). This differs from supervised finetuning (SFT), which learns directly from labeled text sequences. Early follow-up work focused on math and code applications \cite{chen2025towards}, followed by tool-use agents \cite{feng2025retool} where the LLM both calls tools and reasons over their outputs to complete tasks. Compared to earlier approaches to controlling agents such as prompting, scaffolding, and supervised finetuning, RLVR is appealing for its potential to incentivize more general and flexible reasoning and behavior \cite{chu2025sft, guo2025deepseek}.

One major application of tool-use LLMs is knowledge-intensive question-answering. Here, \textit{search agents} can reason about the query and search over knowledge bases (KBs) in an interleaved fashion \cite{yao2023react, trivedi2022interleaving, jin2025search}. RLVR was shown to be effective for training search agents by Search-R1 \cite{jin2025search}, along with many concurrent and follow-up papers \cite{song2025r1, sun2025zerosearch, zheng2025deepresearcher}. However these works emphasize general-knowledge QA that test simple trivia \cite{kwiatkowski2019natural, yang2018hotpotqa, joshi2017triviaqa, ho2020constructing}, and not technical and knowledge-intensive domains like science, engineering, law, and medicine. These require more technical knowledge, reasoning about complex systems, and ability to search technical knowledge bases.

One promising setting for training technical reinforcement learning (RL) search agents is in scientific AI systems \cite{lu2024ai, gao2024empowering}. Scientific research has a huge volume of knowledge in databases and literature \cite{ferguson2014big, delile2024graph}, and traversing that knowledge is an essential part of every stage of the research process \cite{hope2023computational}. The interest in AI search has been established by literature retrieval systems \cite{lala2023paperqa, asai2024openscholar}, and we predict that future complex agent systems for AI research will include modules for searching scientific literature and knowledge bases \cite{lu2024ai}. These search modules will require specialist domain understanding to properly perform query formulation, to reason about retrieved information, and to evaluate the quality of the retrieved information.

In this work, we propose training RL search agents to search and reason over a corpus of research papers to answer scientific questions. We focus on easily-verified factoid questions, for example \textit{What gene is mutated in childhood retinoblastoma?} (Answer \textit{RB1}); such queries are amenable to current RLVR training, while also being useful to practicing scientists \cite{krithara2023bioasq}. Specifically, we first release a corpus and search index of 16 million abstracts from biomedical papers in PubMed. Second, we release a dataset of 60k factoid QAs; the datasets are generated from the Pubmed articles in an LLM workflow, that underwent rigorous quality assurance by biology experts for correctness and relevance to a real scientific search application. The data creation methods are highly scalable, and can be adapted to other domains like materials science or chemistry. Third, and for benchmarks, we reserve 5k samples for testing, and we re-distribute the factoid subset of BioASQ, a small scale but high quality human-created dataset \cite{krithara2023bioasq}.

We train LLM search agents in our environment, showing that current RL training techniques \cite{jin2025search} lead to stronger performance compared to non-RL baselines. However the overall scores remain low, which establish our datasets as challenging for training search systems. We perform quantitative analysis, finding:  general-domain semantic retrievers offer small benefits compared to syntactic retrievers; LLMs without retrievers have non-negligible performance; gains to accuracy with model size are likely due to better parametric knowledge; and paraphrasing in dataset construction adds dataset difficulty. Additionally, our qualitative results show interesting behaviors, specifically simple planning about query rewriting, reasoning about questions before retrieval, and verification when the model already has an initial answer.

In summary, our contributions are: 
- A new environment for training search agents in scientific question answering over papers: specifically a corpus, training datasets, and benchmarks.
- Demonstrating successful RLVR training of search agents over scientific papers, with quantitative and qualitative insights.

\section{Related Work}
We review general-domain search agents, followed by systems for understanding scientific literature.
% We have an extended related work in \Cref{sec:}

\subsection{Search agents}
Search-R1 \cite{jin2025search} and R1-Searcher \cite{song2025r1} were the first open search agents for question answering trained using reinforcement learning with final-answer reward. (Closed systems like OpenAI's o3 \cite{openai2025o3} and Deep Research likely explored this earlier \cite{openai2025deepresearch}). There were many followups exploring, for example, search in web environments \cite{zheng2025deepresearcher, li2025webthinker}, query decomposition \cite{guan2025deeprag}, and simulating the retrieval environment \cite{sun2025zerosearch}. We contribute to this direction by proposing new RL training environments; while prior works emphasize general knowledge QA, we create datasets, evaluations, and a retrieval corpus for training agents to reason over scientific literature. Earlier, search agents (and RAG systems) were supervised with supervised fine-tuning \cite{schick2023toolformer}, few-shot prompting \cite{yao2023react, trivedi2022interleaving}, or prompt optimization \cite{opsahl2024optimizing}; these approaches likely lead to worse generalization \cite{chu2025sft, guo2025deepseek}. Concurrent with recent search agents, many train agents with RL for tool use beyond search engines \cite{feng2025retool, qian2025toolrl}.

\subsection{Search Agents for Scientific QA}
BioASQ \cite{tsatsaronis2015overview, krithara2023bioasq} is an annual challenge run since 2012 for benchmarking semantic indexing and open-domain question-answering for scientific literature -- its popularity reflects the importance of literature understanding tasks for practicing scientists. Their task definitions influence our dataset construction, though a limitation is that their human-generated data is hard to scale. There are many systems for open-domain question-answering over literature, include PaperQA \cite{lala2023paperqa, skarlinski2024language} and OpenScholar \cite{asai2024openscholar}. They  have impressive capabilities, handling large corpora of full-text articles, however the agent behavior is controlled by component scaffolding, prompt engineering, or supervised fine-tuning. Instead, we explore training agents with RL because it promises stronger generalization in the long term \cite{chu2025sft, jin2025search}. 
To make progress in this direction, we focus on \textit{factoid} QA, where answers are easy to unambiguously verify. Note that current RL-trained search agents are designed for factoid QA \cite{jin2025search}, while such questions are useful to applications \cite{krithara2023bioasq}. This motivates us generating new datasets, since prior datasets have binary answers \cite{jin2019pubmedqa, wadden2020fact}, have long-form answers with fuzzy evaluation \cite{asai2024openscholar, lee2023qasa}, or they have smaller scale \cite{skarlinski2024language}.

\section{Methods}

In the following sections, we first describe the training data construction process, then the search corpus and indexing, and finally the RL training algorithm. 

\subsection{Dataset Construction}
\label{sec:data-pipeline}
\begin{figure}[t]
    \centering
    \includegraphics[width=0.49\textwidth,page=1, trim={0 70 1220 0},clip]{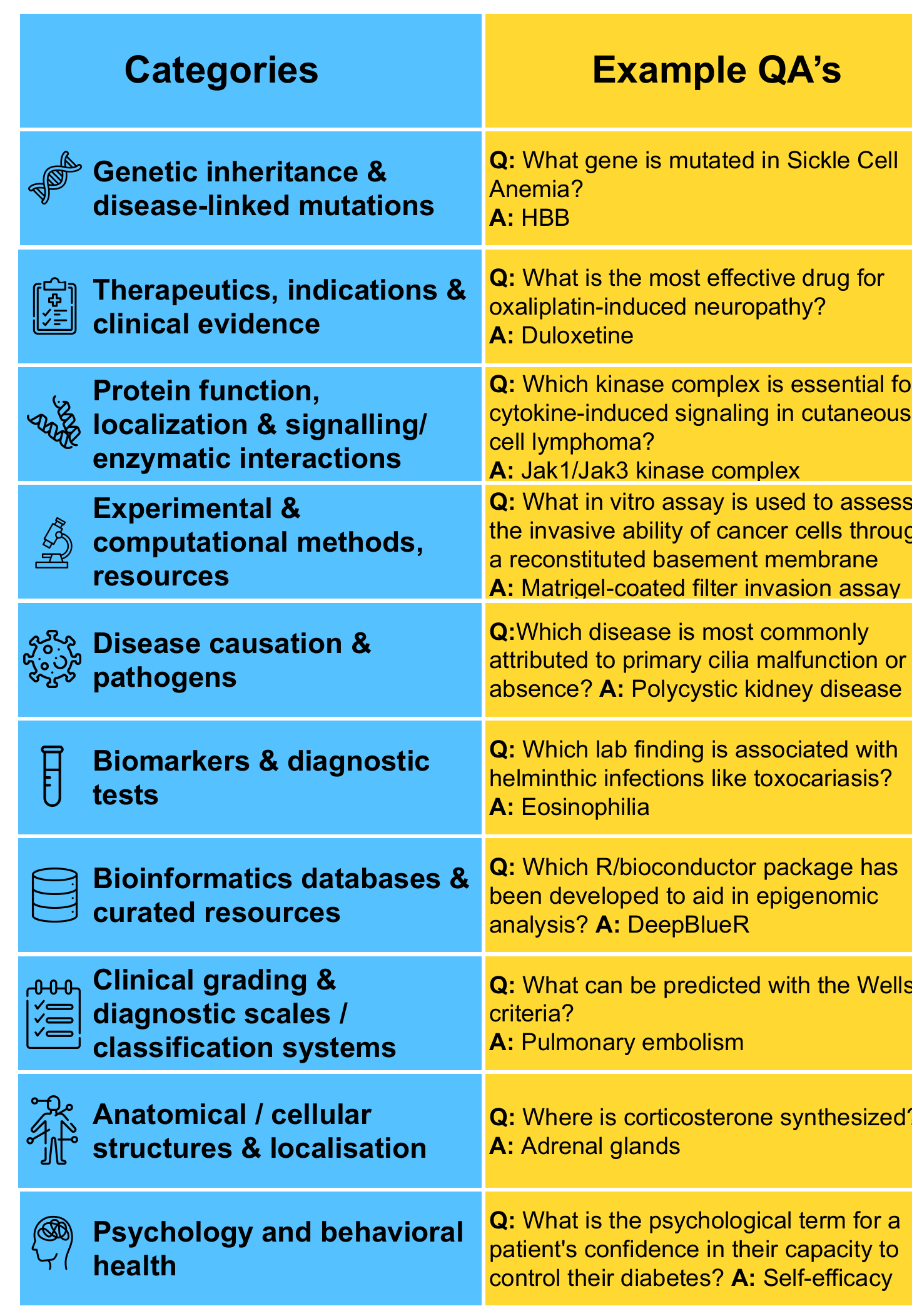}
    \caption{Left: the ten question-answering categories defined with experts. Right: example question-answer pairs, which are sufficient supervision for RLVR training methods.
    }
    \label{fig:categories}
    \vspace{-1.5em}
\end{figure}

\paragraph{Defining Dataset Properties}
The first main goal is that question-answer pairs (QA's) can serve as training data for methods needing \textit{outcome} supervision -- for example, reinforcement learning with verifiable rewards (RLVR) \cite{jin2025search}. Specifically, the answers mut be \textit{verifiable} -- it should be possible for a reward model to judge whether the prediciton matches the ground truth answer without any ambiguity. To satisfy verifiability, we make the following design decisions. QA's are \textit{factoid}, meaning the answer is a single entity; this is similar to the most popular general-knowledge QA datasets studied by search agents \cite{kwiatkowski2019natural, joshi2017triviaqa}. (Alternative and more complex formulations, like `list of entities' have been left to future work). The questions are unambiguous: written so that only a single entity name (or its synonyms) are correct. Then, the reward model is simply checking whether the prediction is equal to the ground truth answer (or its synonyms). We ensure questions have a low `random guessing baseline', because this can lead to incorrect reasoning frequently being rewarded, which is noisy supervision. In particular, we do not allow binary answers (e.g. True or False), and our quality control process ensures that the question text rarely gives a small list of options.  Another property -- implicit in our construction pipeline -- is that questions are single-hop, meaning they can be answered from a single correctly-retrieved document. Since we employ outcome-only reward, we do not require annotations for intermediate reasoning or for retrieved documents.

The second main goal is that QA's should be relevant to real applications: they must be questions that real scientists might ask in their work. To ensure this, our team includes practicing scientists at all stages -- from defining task properties to pipeline construction to verifying the data. Our construction pipeline also take inspiration from the BioASQ project \cite{krithara2023bioasq, nentidis2023overview, tsatsaronis2015overview} -- a challenge for semantic indexing and question-answering (including factoid-QA) over biomedical articles -- that has run since 2015, and garnered significant attention in bioinformatics and NLP. While a limitation of BioASQ is that questions are human-created and therefore difficult to scale, it clearly demonstrates the significant interest in biomedical question answering over scientific papers; this supports our claim that PaperSearchQA is interesting to applications.

\paragraph{Categories for Question-Answering}
To ensure the QA-generation pipeline produces questions that satisfy our key target properties -- unambiguous factoid and relevant to application -- we defined ten target question categories. The categories and examples are shown in \Cref{fig:categories}. 

To develop these, first the human experts on our team performed brainstorming to identify one candidate category set. Next, we sampled 300 questions from the BioASQ database and used LLMs (Claude Opus 4 \cite{anthropic2025_systemcard} and OpenAI o3 \cite{openai2025o3}) to propose two more candidate category sets. Then, the human experts synthesized those into a final list, which required some merging and discarding rare categories. The final category names with examples were used in the data construction pipeline. 

\begin{figure*}[t]
    \centering
    \includegraphics[width=\textwidth,page=2, trim={0 620 375 0},clip]{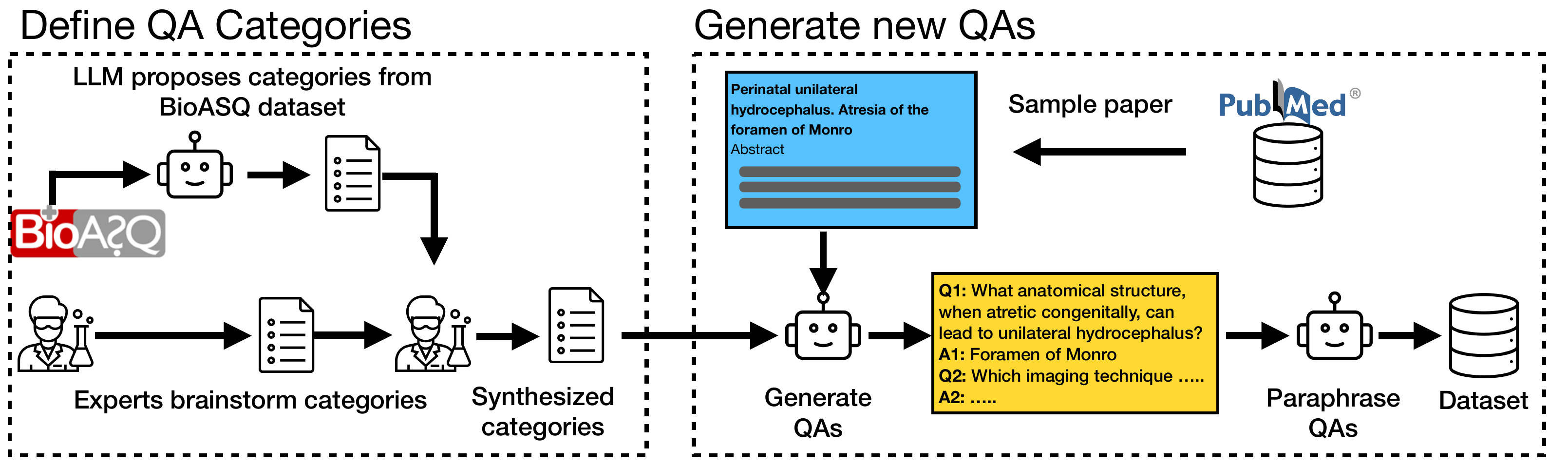}
    \caption{Data generation pipeline process. Left, generating the categories from \Cref{fig:categories}: LLM summarizes categories from human-written questions in BioASQ \cite{krithara2023bioasq}; humans brainstorm categories in parallel; humans synthesize both sources into final categories. Right, QA generation: abstracts from PubMed are sampled and passed to an LLM. The LLM s prompted with categories (and other guidance) to generate QAs. A second LLM paraphrases the QAs to limit exact keyword matching. 
    }
    \label{fig:data-gen-pipeline}
\end{figure*}

\paragraph{Automatic QA Generation Pipeline}
The data generation process (\Cref{fig:data-gen-pipeline}) uses paper abstracts as a knowledge source, which are then mapped to QAs using an LLM workflow. The LLM prompts and pipeline architecture were iteratively designed based on expert review from biomedical scientists. Specifically we generate 200 questions, the expert provides text feedback; the human prompt engineer then modifies the workflow topology or the LLM instructions with metaprompting \cite{schulhoff2024prompt}.

First, the paper abstracts are randomly sampled from the corpus described in \Cref{sec:corpus} -- the same corpus that is searched at inference time. The abstract is passed to an LLM (GPT-4.1 \cite{achiam2023gpt}) with a carefully-designed prompt (see \Cref{sec:appendix-data-pipeline}). This prompt includes the target categories from \Cref{fig:categories}, along with guidance ensuring the questions are suitable for open-domain QA: factoid answers, no acronyms, and no assumed access to the document (and we add an extra filtering step for phrases like `this study'). We found that generating three questions per abstract led to better dataset diversity. 

Since reward models commonly use exact match comparison of prediction and target, we generate synonyms of `golden answers', using GPT-4.1 (prompt in \Cref{sec:appendix-data-pipeline}). Next, we notice that questions often use exact keywords and phrasing found in the abstract, while realistic use-cases would often use synonyms. We therefore sample 50\% of QAs for question rewriting, and use an LLM prompt to `paraphrase' the question with different terminology (prompt in \Cref{sec:appendix-data-pipeline}). Finally, dataset is split into train and test randomly.

All LLM calls were made through OpenRouter. The total cost, including experimentation and final data generation, was estimated at \$600.

\paragraph{Dataset Summary}
The final \Dataset dataset has 54,907 training samples and 5,000 test samples. For question categories \Cref{fig:categories}, the top categories are `Experimental \& computational methods' (27\%) and `Therapeutics, indications \& clinical evidence'. Median question word length is 18 and median answer word length is 2. Each sample is annotated with the Pubmed ID of the source paper, the category, and whether the question was paraphrased to avoid easy keyword matching. It is available on Hugging Face Hub and is released with a CC-BY license.

\subsection{Evaluation dataset: BioASQ}
\label{sec:eval-bioasq}
BioASQ is a popular challenge for biomedical indexing and question answering, where all samples are human-creating \cite{krithara2023bioasq, tsatsaronis2015overview}. Due to it's smaller scale, we propose using it for search agent evaluation, where the search corpus is the same PubMed abstracts from \Cref{sec:corpus}. For convenience, we collect data from all years up to 2025 and redistribute it on Huggingface Hub. Our only addition is to generate synonyms for the answer into the `golden answer' list  (using the same LLM call from our own pipeline) which enables exact-match evaluation metric. It is released under CC-BY-2.5 license. Its `factoid' dataset has 1,609 samples. BioASQ has question categories other than \textit{factoid} -- \textit{yes/no}, \textit{list}, and \textit{summary} -- which we also release, though we do not use it in this paper. 

\subsection{Retrieval Corpus and Index}
\label{sec:corpus}
The search corpus is 16 million PubMed abstracts up to 2025, and was previously distributed by BioASQ \cite{krithara2023bioasq}\footnote{PubMed abstracts originally sourced from \href{https://www.nlm.nih.gov/databases/download.html}{National Library of Medicine}
}. We concatenate the paper title with the abstract text, giving a mean word length of 245. 

We provide BM25 \cite{robertson1994some} and e5 \cite{wang2022text} search indexes. The corpus and index is small enough to hold in memory: the corpus  is 23GB, the BM25 index is 2.6GB, and the e5 index is 93GB. At inference time, the e5 retriever index requires two A100s GPUs (80GB) to avoid memory error at inference.

\subsection{Training Algorithms}
\label{sec:training}
To demonstrate the value of our datasets and retriever, we train search agents using RLVR. 

\paragraph{RLVR for Search Agents}
We follow Search-R1 \cite{jin2025search}, which uses reinforcement learning with verifiable rewards (RLVR). 

We provide a minimal system prompt (\Cref{sec:appendix-system-prompt}), which introduces the question-answering task, instructing the model to leverage reasoning tokens inside \texttt{<think>} tokens and to give the final answer inside \texttt{<answer>} tokens. The prompt then describes usage of search: by wrapping queries in \texttt{<query>} tokens. When a query is found, the system stops generation, extracts the query, and retrieves the top $k$ documents. It then appends the documents to the reasoning trace, and then continues token generation. Crucially, this system prompt provides minimal specific guidance about how to perform reasoning and query rewriting -- this allows behaviors to be learned in RL training in a manner that (hopefully) is more flexible and general \cite{chu2025sft}.

In training, the search agent performs rollouts of token generation and search. The final answer is extracted and we compute a very simple reward: 1 if the prediction matches any of the target answers, and 0 otherwise. Reward is applied to all LLM-generated tokens uniformly, except for the retrieved tokens that are masked out during gradient computation. More formally (as in Search-R1 \cite{jin2025search}) we learn the weights for the policy LLM, $\pi_\theta$, conditioned on a retrieval engine $\mathcal{R}$ using a QA dataset, $\mathcal{D}$:

\begin{align*}
    \max_{\pi_\theta}\  &\mathbb{E}_{x \sim \mathcal{D}, y \sim \pi_{\theta}(\cdot \mid x; \mathcal{R})} \left[ r_{\phi}(x, y) \right] \\
    &- \beta \mathbb{D}_{\text{KL}} \left[ \pi_{\theta}(y \mid x; \mathcal{R}) \,||\, \pi_{\text{ref}}(y \mid x; \mathcal{R}) \right]
\end{align*}

In the first term, the LLM generates tokens, $y$ from the question $x$, conditioned on a retriever: $y\sim\pi_\theta(\cdot\|R)$. The reward model, $r_\phi(x,y)$, extracts the answer from the sequence and compares against ground truth. In the second term, the policy LLM, $\pi_\theta$, is discouraged from diverging too far from a reference LLM $\pi_\theta$, which is the LLM's initial state. We use Group Relative Policy Optimization (GRPO) to optimize the LLM based on the samples; further details in \Cref{sec:appendix-training}.

\section{Results}
To demonstrate the utility of our dataset, corpus, and benchmarks, we train the LLM with reinforcement learning with verifiable rewards (RLVR). Our experiments show that RLVR training improves performance on scientific paper question-answering evaluations. We also provide further quantitative and qualitative analysis.

\subsection{Experiment details}
\paragraph{Baseline methods}
We build our dataset to facilitate training with RLVR, which supervises only the final answer and thus, promises stronger generalization compared to methods with heavy scaffolding or with reasoning SFT  \cite{guo2025deepseek, chu2025sft}. To validate this strategy, we compare RLVR training to baseline LLM training approaches that impose few assumptions: direct LLM inference, chain-of-thought prompting \cite{wei2022chain, kojima2022large}, retrieval augmented generation (RAG) \cite{lewis2020retrieval}, Search-o1 \cite{li2025search}, and PaperQA2 with the same retriever as other methods
\cite{skarlinski2024language}. 
For a fair comparison, we apply the same base LLM that was used in agent training.

\paragraph{Search-R1 RLVR Training}
We follow the Search-R1 training setup \cite{jin2025search} as described in \cref{sec:training}, and experiment with two retrievers: bm25 and e5. we use eight a100s (80gb) for training, using grpo for 150 steps (runtime: ca. 30\ hrs). we have batch size 512 and minibatch size 256 for two gradient updates per batch (full configuration is in the code). the training framework is \emph{verl} \cite{sheng2024hybridflow}. the base llms are qwen2.5 3b and 7b, and we experiment with both \emph{base} and \emph{instruct} \cite{qwen2.5, qwen2}. 

\paragraph{Evaluation}
We evaluate with the test set of \Dataset, and the BioASQ-factoid benchmark \cite{krithara2023bioasq} version that we release (\Cref{sec:eval-bioasq}). The evaluation metric is the same as the RL training reward term: the prediction must exactly match one of the ground-truth answers, which are all synonyms (for example target answer `APOC3' has synonyms  `apolipoprotein C-III', `apoC-III', `apoCIII',
`apolipoprotein C-III', `apolipoprotein C3', among others). The matching function includes `normalization': conversion to lower case, stripping leading and trailing whitespace, removing articles like `a' and `the'.

\begin{table}[]
\label{tab:results}
\centering
% \begin{tabular}{@{}lrr@{}}
% \resizebox{\columnwidth}{!}{%
{
\begin{tabular}{@{}lrr@{}}
\toprule
 & \multicolumn{1}{l}{\textbf{PaperSearchQA}} & \multicolumn{1}{l}{\textbf{BioASQ}} \\ \midrule
\multicolumn{3}{l}{\textbf{Qwen2.5-3b-Instruct}} \\
Direct & 16.7 & 15.8 \\
CoT & 20.3 & 16.5 \\
RAG & 32.0 & 30.0 \\
Search-o1 & 30.8 & 29.4 \\
PaperQA2 & 32.4 & 33.1 \\
SearchR1 & 41.6 & 35.5 \\ \midrule
\multicolumn{3}{l}{\textbf{Qwen2.5-7b-Instruct}} \\
Direct & 27.5 & 24.9 \\
CoT & 29.7 & 23.4 \\
RAG & 36.5 & 29.7 \\
Search-o1 & 36.5 & 31.5 \\
PaperQA2 & 37.1 & 32.8 \\
SearchR1 & 51.0 & 44.8
\end{tabular}
}
\caption{Main results of baselines vs Search-R1 training \cite{jin2025search} that uses RLVR. The metric is accuracy, where `correct' is exact match of prediction to target (or a synonym for the target). PaperSearchQA is the test set of our dataset, while BioASQ is a human-created evaluation. The RAG and Search-R1 systems used BM25 retrieval, and we compare to e5 retriever in the text.}
\end{table}

\begin{figure*}[t]
    \centering
    \includegraphics[width=\textwidth,page=4, trim={0 360 160 0},clip]{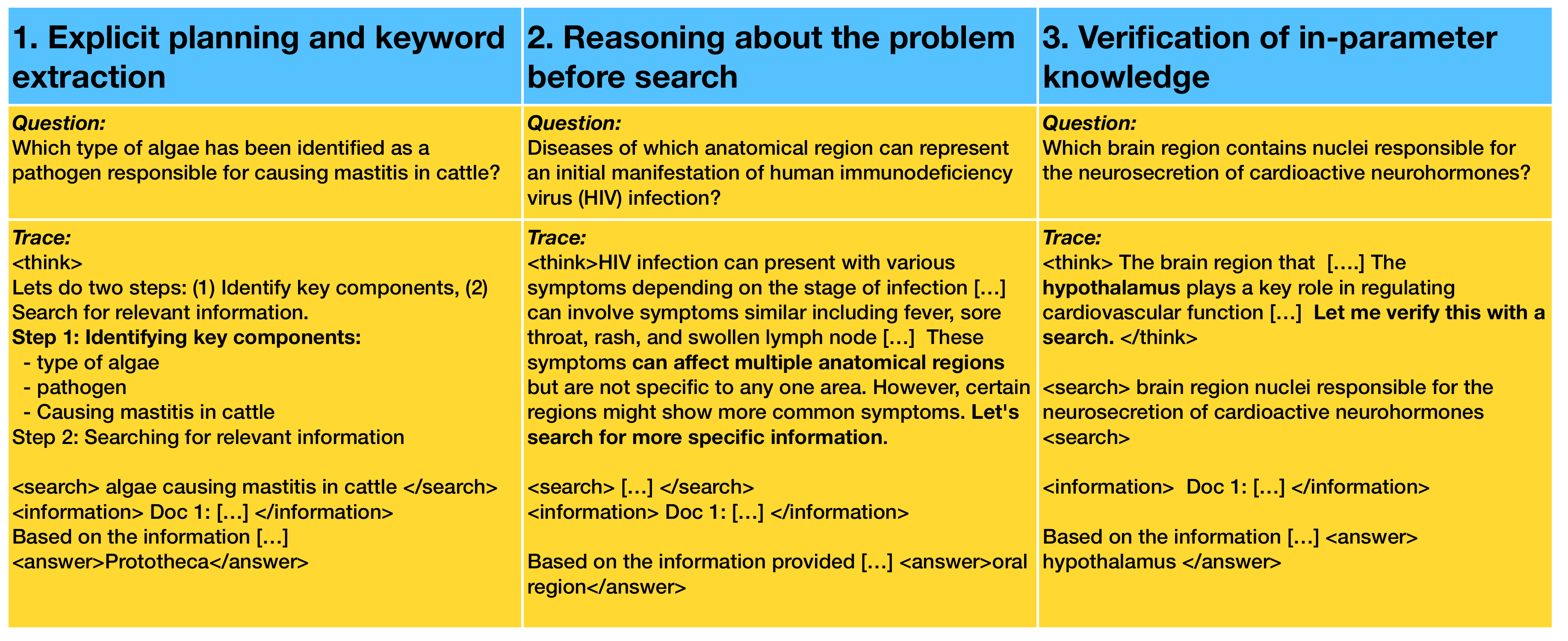}
    \caption{Three interesting behaviours that we observe in search agent traces. We bold some words for emphasis. Since traces are long, we abbreviate them, as indicated by `[...]'. These are discussed further in \Cref{sec:results-qualitative}.
    }
    \label{fig:qualitative-traces}
\end{figure*}

\subsection{Quantitative Results}
The main results are in \Cref{tab:results}, showing accuracy on the target benchmarks, and where the base model was \emph{Qwen-instruct} (both, 3B and 7B variants). Training with RLVR (specifically using Search-R1 \cite{jin2025search}) clearly leads to the strongest results. For the 3B LLMs, RL improves over RAG by 9.6 and 5.5 points for PaperSearchQA and BioASQ respectively. For 7B models, the difference is 14.5 and 9.3. RAG outperforms the retrieval-free methods by 17 points on average. Chain-of-thought prompting outperforms direct inference by only 1.2 points on average.

\Cref{tab:results-per-category} shows per-category results for all models. The easiest overall categories are `Biomarkers \& diagnostics' and `Protein function \& signalling', while `Genetic mutations' is the most challenging.

We perform further quantitative analysis and share these additional findings:

\textit{Semantic retrieval gives little benefit over syntactic retrieval} For both RAG and RL training, we experimented with the BM25 syntactic retriever, and the e5 semantic retriever. While the semantic retriever should help search where exact keywords differ, the performance benefit was minor -- within 2 points in all experiments. One possibility is that, even when paraphrasing questions, it must include certain technical keywords, which makes retrieval easier. Another possibility is that the e5 retriever under-performs for scientific domains (which involve highly technical terminology), thus removing the benefit of semantic retrieval.

\textit{LLMs encode scientific knowledge} The retrieval-free baseline scores (from \Cref{tab:results}) are reasonably high, and scale with model size. For example on \Dataset\ they score 20.3 and 29.7 for 3B and 7B models. This is probably explained by the fact that PubMed abstracts are easy to download, and so they likely appear in pretraining mixtures. Despite this data (probably) being seen by the model, memorization is far from perfect, so retrieval remains necessary.

\textit{Superior performance with model size is likely due to knowledge} Averaged across benchmarks, Search-R1 outperforms CoT by 20.2 points for the 3B model and 21.4 for the 7B model. This suggests that the performance gain is due to improved parametric knowledge, and not due to superior capabilities in query formulation or  comprehension. 

\textit{Paraphrasing in data construction is beneficial} In dataset construction, we observed that LLM-generated questions would often mirror keywords or phrasing from the source document in \Cref{sec:data-pipeline}, and so we added a paraphrasing step to 50\% of the QAs, allowing to compare non-paraphrases and paraphrased QAs. For SearchR1 trained on \Dataset, non-paraphrased questions scored 57.2 while paraphrased questions scored 44.9, highlighting the importance of paraphrasing for sustaining question difficulty. 

\textit{Training dynamics are similar to general-domain QA training environments} The Search-R1 study \cite{jin2025search} observed certain dynamics that we also observe. Specifically, we observed small performance difference between base and instruct models, albeit the base model required more training time to converge. We also found that training with GRPO was unstable, and reward would collapse to zero for some training runs -- the base (non-instruct) models were generally more stable.

\subsection{Qualitative Results}
\label{sec:results-qualitative}
To better understand the system performance, we manually reviewed the reasoning traces for models at multiple stages in training. We highlight three prevalent patterns in \Cref{fig:qualitative-traces}. The format of the traces includes reasoning inside `<think>' tokens and the final answer is in `<answer>' tokens. To perform retrieval, the LLM outputs text in `<search>` tags; the retrieved documents are dumped into the trace inside `<information>`. 

\textit{Behavior 1 -- explicit planning and keyword extraction.} We find this pattern to be very common in later training. The model follows a clear and simple strategy common in RAG with rewriting: extracting the keywords for search and then combining them into a search query. After performing search, the LLM summarizes the final conclusion.

\textit{Behavior 2 -- reasoning before search.} Here, the LLM reasons about the question using only its parametric knowledge before performing any search. In the example problem, it observes that disease symptoms vary based on stage, and suggests symptoms from its own parametric knowledge. The trace acknowledges that it does not have the answer, and performs search. After viewing the retrieved information, the presence of earlier reasoning tokens may impact the final answer.

\textit{Behavior 3 -- verification of in-parameter knowledge.} The LLMs have sufficient knowledge to answer between 15\% and 30\% of questions (\Cref{tab:results}), so how does the agent behave when it already knows the answer? We find that it performs search anyway, but in the reasoning trace it will state its initial answer, and explicitly declare that it is doing further verification. Verification is generally good, since the LLM can gather more evidence for a reliable answer. More sophisticated systems however should only search when not confident in its initial answer.

\textit{Agent behavior becomes less varied  with more training} With more training, behavior 1 becomes much more common. We suspect this is due to lack of training data diversity -- \Dataset only includes factoid-QA, and so this learned strategy is effective for most samples. Future systems  trained on more QA types and elicit more varied behavior.
% for models initialized from both \emph{base} and \emph{instruct}

\textit{Very little reasoning after viewing documents} After adding retrieved documents, the LLM tends to answer immediately, without explicit reasoning about document contents. This could be explained by comprehension being simpler with factoid-QA; it is also possible that RL training led to better comprehension due to parameter weight updates.

\section{Discussion}
\label{sec:discussion-and-conclusions}
We show that search agents can be trained using RL to perform question-answering by reasoning and gathering knowledge from scientific papers, a crucial intellectual part of science \cite{tsatsaronis2015overview, hope2023computational}. Search agents -- and more generally RL-trained tool-use agents -- are rapidly advancing in general-domain AI. Our aim in designing the training datasets, benchmarks, and corpus was to ensure compatibility with these methods. We hope that advances to general-domain agents -- both in open research and in private labs -- will translate to stronger capabilities in scientific literature understanding by leveraging our artifacts and others from the AI for science community. 

While our datasets represent progress for scientific search agents, the scope is limited to only single-hop factoid-QA and simple retrieval over a database of abstracts -- there is huge potential for further work. Interesting directions include factoid-QA designed to be multihop \cite{kim2025biohopr},  answers with list-of-entities, and questions requiring extended answers or summaries \cite{krithara2023bioasq, asai2024openscholar}; these can require more complex agent planning behavior and fuzzy reward models. Even more ambitiously, future work could aim to resolve questions with conflicting evidence, like in critical literature review, \cite{lieberum2025large, polzak2025can, clark2025generative}. Moreover, future datasets should consider that recent results in RLVR for (non-tool-use) LLMs are leveraging LLM-as-a-judge for reward modeling \cite{su2025crossing, gunjal2025rubrics}. Meanwhile, other tool-use and search agent works consider text and images, which is relevant to scientific papers as well \cite{wu2025mmsearch, wang2025vrag}. 

Other research directions are more specific to literature understanding applications. Agents could be equipped with tools and metadata that would be used by real scientists in their work, for example citation traversal and source reliability metrics. For example, one could implement a scoring on to what extent the conclusions extracted from a scientific article are supported by the figure images / data presented in the article -- an assessment that is typically made by scientists when they deeply review literature. This could aid in valuing contradicting or diverging scientific results for a reply. Such metrics could be provided in the output, which could contain multiple answers with scores.

On a final note, our data generation pipeline is quite general -- it could be adapted to generate QA datasets in other domains like chemistry, materials science, and computer science.

\section{Conclusion}
AI  holds great potential to transform science. One exciting cluster of methods are LLM agents or multi-agent systems -- sometimes called AI Scientists \cite{ gao2024empowering, lu2024ai, gottweis2025towards, huang2025biomni, hope2023computational}. This research program anticipates agents becoming more and more autonomous -- first by performing well-defined tasks like data analysis and experimental execution (e.g., \cite{huang2025biomni}) -- and later performing more open-ended tasks \cite{hughes2024open} like planning new experiments and even forming new hypotheses. But scientific fields are deeply knowledge-intensive: scientific discovery requires recalling, retrieving, and evaluating arcane information in the massive corpus of human knowledge. We therefore claim that future AI Scientist systems will require the capability of knowledge intensive search. Literature understanding is therefore fundamental to AI systems in science, and we believe that RL training of search agents -- like in this paper -- is an essential approach.

\clearpage

\section{Limitations}

First, the data generation pipeline is automatic and uses LLMs, which could lead to factually incorrect QAs. One source of risk is LLM hallucination, though the risk is small since each prompt has a smaller context, and we use strong LLMs (GPT-4.1). 

Another risk from our data generation pipeline is that it is challenging to infer a `general QA' from a single specific abstract. For example, an abstract might claim ``mutation in gene X correlates with disease Y'', and our pipeline might derive the question ``what gene mutation is correlated with disease A?''. But since we only have one abstract in context, we cannot be sure that `gene mutation X' is the only answer -- some other abstract might report that `gene mutation Y' also correlates with the disease. In designing our data generation pipeline, expert review found such cases to be rare, and so we did not design complex mitigations.  (It is possible that some such questions were avoided due to the parametric knowledge in the LLM generating the questions -- GPT-4.1 -- which is a more capable model than the smaller models used in these experiments). Future work that follow our data generation methodology could apply mitigations if needed. For example, if human review finds the issue prevalent for certain question categories, then that category could be excluded. Or, a workflow could be designed to retrieve all relevant papers to check for conflicts (which would be allowed a large retrieval budget). 

In terms of scope, this is a first study in using RLVR to train search agents, so we restricted it to factoid QA. While this is a similar restriction to other early search agent papers, it represents only one of the possible question types important for real applications -- we discuss future directions in \Cref{sec:discussion-and-conclusions}. Likewise, our dataset covers scientific papers in biology \& medicine, but not other domains commonly studied in AI for science like chemistry, materials science, computer science. However most AI for science papers have a similar limitation because significant domain expertise is required, making highly general studies challenging \cite{mirza2025framework, burgess2025microvqa, tang2025matterchat}.

Still on scope, another limitation is that we tackle text-only problems, however it is scientific reasoning obviously goes beyond text, for example to consider images \cite{yue2024mmmu, burgess2025microvqa} and more general data types \cite{huang2025biomni}. Future work could use data sources like BIOMEDICA \cite{lozano2025biomedica} that include paper figures for PubMed open-access articles. 

This synthetic data generation procedure requires access to research articles, which are often protected by copyright; the field should consider approaches similar to \cite{schuhmann2025project} to overcome this. Future systems could also better leverage science-specific retrieval systems \cite{li2025r2med, asai2024openscholar}.

While the study provides resources towards building useful search agents for scientific practitioners, the derived agent system is a research prototype and is not suitable for real-world use. Apart from having a too-restricted scope, it has not undergone thorough evaluation needed for real-world deployment. 

% \FloatBarrier
% \clearpage

{
    \small
    \bibliography{custom}
}

\clearpage
\appendix
\clearpage
\setcounter{page}{1}
% \onecolumn

\section*{Acknowledgments}
We gratefully acknowledge NVIDIA's Academic Grant Program for providing cloud GPU resources used in this research.

\section{Dataset and Code Availability}
\paragraph{Accessing data}
We release all artifacts on the Huggingface Hub at \href{https://huggingface.co/collections/jmhb/papersearchqa}{https://huggingface.co/collections/jmhb/papersearchqa}.

\paragraph{Accessing Code} The code at \href{https://github.com/jmhb0/PaperSearchQA}{https://github.com/jmhb0/PaperSearchQA} is for 

\paragraph{Licenses}
Our dataset, \Dataset, is released under a fully open license CC-BY-4.0, permitting redistribution, remixing, and commercial use. The data is derived from PubMed abstracts that are available for bulk download under NLM's Terms and Conditions\footnote{\href{https://www.nlm.nih.gov/databases/download.html}{https://www.nlm.nih.gov/databases/download.html}}. The search corpus and the BioASQ evaluation set are sourced from the BioASQ project \cite{krithara2023bioasq, tsatsaronis2015overview}, and inherit their CC-BY-2.5 license.

\section{Ethical considerations}
This paper advances systems that answer scientific questions from literature, but this presents some risks:
\begin{itemize}
    \item Agents may retrieve and amplify outdated, retracted, or flawed studies without quality assessment mechanisms.
    \item Papers retrieved by the agent may have some selection bias that is poorly understood, thus impacting papers seen by scientists.
    \item Hallucinations in LLM outputs and incorrect QA responses may harm scientific practice.
    \item Our dataset was generated in an automated pipeline, which may have introduced errors. 
\end{itemize}
Future deployments should consider uncertainty quantification, and source quality indicators. More broadly, the scientific community must develop its own standards for the appropriate use of LLM tools that consider these risks.

\section{Statement on use of LLMs}
LLMs were used at many points in the project. Other than what is discussed in the main paper, we had these use cases:
\begin{itemize}
    \item In project conception: brainstorming ideas; giving feedback and criticism on project plans; searching related work; summarizing and answering questions about specific related work.
    \item In project execution: LLMs for code generation in the Cursor IDE. 
    \item Paper writing: rephrasing individual sentences.
\end{itemize}

% \section{Dataset information}
% The sumary stats for the categories. 

\section{Further explanation of reinforcement learning with verifiable rewards (RLVR)}
\label{app:rlvr}

RLVR \cite{lambert2024t} is a post-training procedure in which a language model is optimized only from whether its \emph{final} output can be automatically verified as correct. At a high level, the model proposes a solution to a task, a separate verifier evaluates that solution, and the model is updated to make successful solutions more likely in the future.

\paragraph{Single-turn RLVR.}
Much of the earliest RLVR work uses a single-turn setting, where the model answers in one shot without explicit tool calls or multiple interaction steps. Given a query $x$, the model samples a final answer $y \sim \pi_\theta(\cdot \mid x)$, such as a free-form solution to a math problem or a code snippet. A verifier $V$ then returns a (typically scalar) reward
\[
r = V(x, y),
\]
for example by exact-match against a reference answer, a numerical tolerance check, or running unit tests on the generated code. In many RLVR setups, $r$ is binary ($r \in \{0,1\}$) to indicate pass/fail, but the formulation also allows graded or shaped rewards (e.g., partial credit or the proportion of tests passed).

In this setting, the RLVR objective is
\[
J(\theta) = \mathbb{E}_{x \sim \mathcal{D},\, y \sim \pi_\theta(\cdot \mid x)}[\,r\,],
\]
which says: sample questions $x$ from a data distribution $\mathcal{D}$, sample answers $y$ from the model, and maximize the expected reward returned by the verifier. This captures the basic “generate–verify–reinforce” loop used in early RLVR for math and code.

\paragraph{Multi-step RLVR with trajectories.}
For agents that call tools or take multiple reasoning steps, it is helpful to view RLVR in a more general trajectory form. Given a query $x$, the model interacts with its environment to produce a trajectory
\[
\tau = (o_0, a_0, o_1, a_1, \dots, o_T, y),
\]
where $o_t$ are observations (e.g., tool outputs or intermediate text), $a_t$ are actions (e.g., tool calls or tokens), and $y$ is the final answer returned to the user. The single-turn setting above is a special case where there are no intermediate observations or actions and $\tau$ consists only of the generated answer $y$.

A verifier $V$ now maps $(x, \tau)$ or $(x, y)$ to a scalar reward
\[
r = V(x, \tau).
\]
The verifier can use only the final answer (e.g., exact match or unit tests) or the whole interaction (e.g., whether a sequence of tool calls satisfies some constraints). Let $\pi_\theta(\tau \mid x)$ denote the model’s policy over trajectories; RLVR then maximizes
\[
J(\theta) = \mathbb{E}_{x \sim \mathcal{D},\, \tau \sim \pi_\theta(\cdot \mid x)}[\,r\,].
\]
When $r$ is binary, this reduces to maximizing the probability that the verifier accepts the trajectory, but the same objective accommodates more general reward shapes.

In practice, $J(\theta)$ is maximized using policy-gradient methods. In our experiments we use Group Relative Policy Optimization (GRPO; see \cref{sec:appendix-training}), a variant that uses group-normalized advantages, clipping, and a KL penalty to a reference policy. For intuition, one can view these methods as refinements of the basic REINFORCE estimator
\[
\nabla_\theta J(\theta) \approx \mathbb{E}\Big[(r - b)\sum_{t=0}^{T} \nabla_\theta \log \pi_\theta(a_t \mid h_t)\Big],
\]
where $b$ is a baseline that reduces variance. High-reward trajectories increase the log-probabilities of their actions, while low-reward trajectories decrease them.

\paragraph{Relation to SFT and RLHF.}
RLVR differs from supervised finetuning (SFT) and RLHF in two key ways. First, RLVR uses only verifiable success or failure of the \emph{final} output as a learning signal; there are no human-written labels on intermediate steps and no preference scores over partial generations. Second, credit assignment is purely outcome-based: all intermediate reasoning, tool calls, and textual tokens are reinforced or discouraged according to the reward returned by the verifier. This makes RLVR particularly natural for tasks where correctness can be automatically judged but good intermediate supervision is expensive or unavailable.

% \subsection{More general background} Tool use: 
% \cite{feng2025retool,bespoke_improving_multi_turn_tool_use, qian2025toolrl}

\section{Data construction pipeline}
\label{sec:appendix-data-pipeline}
We show the prompts here. The prompts are long, so for more readablity, refer to the code at \verb|data_gen/generate_questions_from_abstracts.py|

Here is the main data generation prompt mapping an abstract to QAs.
\begin{lstlisting}[basicstyle=\ttfamily\small, breaklines=true, breakatwhitespace=true]
BACKGROUND
You are a domain-expert biomedical NLP assistant.
You are helping me to create an open-domain QA dataset. 
The downstream task will read a query and require an agent to search over Pubmed abstracts

--------
YOUR TASK 
I will provide you with title and abstract of a Pubmed article. 
Your task is to create 3 new question-answer pairs. 

--------
TYPES OF QUESTIONS
The questions should be 'factoid based'. 
The answer should be a simple entity. 
It should not be ambiguous.
Don't be pretentious. 

--------
IMPORTANT NOTES
The question-answer pair will be used to evaluation question-answering systems with retrieval. Ths means the target system does not know which paper the question was sourced from. So an inappropriate question would be "What technology is used in this study to ...". or "what type of treatment is assessed in this study?" (where the study name is not specifified).
If the question contains acronyms that are not well known, then explain the acronym.

--------
EXAMPLE CATEGORIES 
Below are sample categories with sample questions. 

Category: 1 - Genetic inheritance & disease-linked mutations
question: What gene is mutated in Sickle Cell Anemia?
answer: HBB
question: Which ultraconserved element is associated with Embryonic Stem Cells (ESC) self-renewal?
answer: T-UCstem1
question: Is Huntington's disease caused by a dominate or recessive gene?
answer: dominant

Category: 2 - Therapeutics, indications & clinical evidence
question: What is the most effective drug for oxaliplatin-induced neuropathy?
answer: Duloxetine
question: Which cancer is the BCG vaccine used for?
answer: Non-muscle Invasive Bladder Cancer
question: How many injections of CLS-TA did the patients participating in the PEACHTREE trial receive?
answer: two

Category: 3 - Protein function, localization & signalling/enzymatic interactions
question: Which histone mark distinguishes active from inactive enhancers?
answer: H3K27ac
question: Which component of the Influenza A Virus affects mRNA transcription termination?
answer: NS1
question: Which is the main calcium binding protein of the sarcoplasmic reticulum?
answer: Calsequestrin

Category: 4 - Experimental & computational methods, resources & acronyms
question: Which algorithm has been proposed for efficient storage of WGS variant calls?
answer: SeqArray
question: What is an acceptable sequence coverage(depth) required for human whole-exome sequencing?
answer: 30x-60x

Category: 5 - Disease causation & pathogens
question: Which is the most common disease attributed to malfunction or absence of primary cilia?
answer: ['Polycystic kidney disease', 'PKD']
question: What organism causes scarlet fever also known as scarletina?
answer: ['Group A Streptococcus', 'Streptococcus pyogenes']
question: The pathogen Fusarium graminearum affects what type of plant species?
answer: cereal crops

Category: 6 - Biomarkers & diagnostic tests
question: Salivary Cortisol is a biomarker for what disease/syndrome/condition?
answer: stress
question: What is the gold standard for a diagnosis of narcolepsy?
answer: ['Sleep study', 'overnight polysomnography']

Category: 7 - Bioinformatics databases & curated resources
question: Which R/bioconductor package has been developed to aid in epigenomic analysis?
answer: DeepBlueR
question: Which database associates human noncoding SNPs with their three-dimensional interacting genes?
answer: 3DSNP
question: What is the RESID database?
question: Which is the literature-based database of phenotypes?
answer: PheneBank

Category: 8 - Clinical grading & diagnostic scales / classification systems
question: What can be predicted with the Wells criteria?
answer: pulmonary embolism
question: Symptoms of which disorder are evaluated with the Davidson Trauma Scale?
answer: ['post-traumatic stress disorder', 'PTSD']
question: Which value of nuchal translucency thickness is set as the threshold for high-risk for Down Syndrome?
answer: 3mm

Category: 9 - Anatomical / cellular structures & localisation
question: Where is corticosterone synthesized?
answer: Adrenal glands
question: Which is the chromosome area that the human gene coding for the dopamine transporter (DAT1) is located to?
answer: 5p15.3
question: Where is the respirasome located?
answer: inner mitochondrial membrane

Category: 10 - Psychology and behavioral health
Question: Which psychomotor domain showed a significant difference between institutionalized and non-institutionalized sheltered children and adolescents?
Answer: Body awareness
Question: What ethical principle justifies actions that have both good and harmful effects, as long as the harm is not intended but only foreseen?
Answer: Rule of Double Effect
Questions: What psychological process during an incubation period is associated with enhanced creative problem solving?
Answer: Mind-wandering

--------

OUTPUT FORMAT
A single QA has tags `<question>...</question>`, answer inside `<answer>...</answer>`. 
If the QA corresponds to one of the above categories put its number in <cat_num>...</cat_num> and category description in <cat>...</cat>. 
Each QA should exist in its own tag <qa>...</qa>

Therefore the first 2 questions would be:
<qas>
   <qa> <question> ... </question>
      <answer> ... </answer>
      <cat_num> ... </cat_num>
      <cat> ... </cat>
   </qa>
   <qa> 
       .....
   </qa>
   ...
</qas>

--------
TITLE AND ABSTRACT
{title_abstract}
"""
\end{lstlisting}

And here is the prompt for generating `golden answers' or synonyms to the ground truth answer.
\begin{lstlisting}[basicstyle=\ttfamily\small, breaklines=true, breakatwhitespace=true]
You are given a question that was written using a particular document as its main source. Your task is to rewrite the question so that it retains the original meaning and would result in the same correct answer, but uses different wording and phrasing. Important constraints:
Do not broaden or narrow the scope of the question.
Do not introduce ambiguity or alter clinical/technical context.
Make sure the correct answer remains exactly the same.
Your goal is to change the surface wording so that simple bag-of-words search (like BM25) may not easily match the original document, while an expert human or strong language model could still answer correctly.
Avoid copying any significant phrase (three or more words in sequence) from the original question.

Example: 
- Original: What congenital abnormality can cause unilateral hydrocephalus in the perinatal period? 
- Edited: Which birth defect present during the perinatal stage may result in hydrocephalus affecting only one side of the brain?

Output should be in tags like <question> ... </question>

Question: {question}
Answer: {answer}
\end{lstlisting}

\section{System prompt for Search-R1 LLM training}
\label{sec:appendix-system-prompt}
The LLM system prompt provides basic guidance about what tools are available, as well as guidance about putting the final answer in tags.

\begin{lstlisting}[basicstyle=\ttfamily\small, breaklines=true, breakatwhitespace=true]
Answer the given question. You must conduct reasoning inside <think> and </think> first every time you get new information. After reasoning, if you find you lack some knowledge, you can call a search engine by <search> query </search> and it will return the top searched results between <information> and </information>. You can search as many times as your want. If you find no further external knowledge needed, you can directly provide the answer inside <answer> and </answer>, without detailed illustrations. For example, <answer> Beijing </answer>. Question: {question}\n
\end{lstlisting}

For baseline experiments we apply the same formatting instruction.

\section{Results: per-category performance}
Since PaperQA2 has per-category labels (\cref{fig:categories}), we report the main results split by these category values. The main results are in \cref{tab:results-per-category}.

% Please add the following required packages to your document preamble:
% \usepackage{booktabs}
\begin{table*}[]
\resizebox{\textwidth}{!}{%

\begin{tabular}{@{}l|r|rrrrr|rrrrrr@{}}
\toprule
 & \multicolumn{1}{l|}{\textbf{Data portion}} & \multicolumn{1}{l}{\textbf{3b models}} & \multicolumn{1}{l}{} & \multicolumn{1}{l}{} & \multicolumn{1}{l}{} & \multicolumn{1}{l|}{} & \multicolumn{1}{l}{\textbf{7b models}} & \multicolumn{1}{l}{} & \multicolumn{1}{l}{} & \multicolumn{1}{l}{} & \multicolumn{1}{l}{} \\ \midrule
 & \multicolumn{1}{l|}{} & \multicolumn{1}{l}{Direct} & \multicolumn{1}{l}{CoT} & \multicolumn{1}{l}{RAG} & \multicolumn{1}{l}{PaperQA2} & \multicolumn{1}{l|}{Search-R1} & \multicolumn{1}{l}{Direct} & \multicolumn{1}{l}{CoT} & \multicolumn{1}{l}{RAG} & \multicolumn{1}{l}{PaperQA2} & \multicolumn{1}{l}{Search-R1} \\
Genetic mutations & 3.6 & 12 & 17 & 40 & 20 & 27 & 18 & 18 & 45 & 19 & 26 \\
Therapeutics \& clinical evidence & 17 & 17 & 23 & 31 & 28 & 38 & 27 & 32 & 37 & 32 & 46 \\
Protein function \& signalling & 12.36 & 15 & 20 & 39 & 32 & 44 & 28 & 29 & 46 & 37 & 53 \\
Methods \& resources & 26.36 & 14 & 16 & 26 & 25 & 35 & 26 & 25 & 30 & 27 & 37 \\
Disease causation \& pathogens & 12.96 & 24 & 27 & 38 & 33 & 39 & 34 & 38 & 43 & 36 & 52 \\
Biomarkers \& diagnostics & 10.38 & 20 & 19 & 26 & 34 & 46 & 29 & 32 & 30 & 40 & 56 \\
Bioinformatics databases & 0.16 & 13 & 25 & 13 & 100 & 100 & 13 & 13 & 13 & 100 & 100 \\
Clinical scales \& classifications & 2.82 & 16 & 16 & 26 & 25 & 34 & 23 & 26 & 28 & 34 & 50 \\
Anatomy \& cellular localisation & 8.74 & 13 & 22 & 39 & 24 & 32 & 27 & 30 & 42 & 28 & 37 \\
Psychology \& behavioural health & 3.4 & 16 & 19 & 28 & 26 & 33 & 27 & 31 & 31 & 30 & 39 \\ \bottomrule
\end{tabular}
}
\caption{Main results of baselines vs Search-R1 training \cite{jin2025search} that uses RLVR. Unlike the table in the main results, we show the per-category scores, where the categories are defined in \cref{fig:categories}.
}
\label{tab:results-per-category}
\end{table*}

\section{Results: a note on PaperQA baseline}
For PaperQA baselines we used the official codebase (\href{https://github.com/Future-House/paper-qa}{https://github.com/Future-House/paper-qa}) and then for fair comparison with other methods, we matched the model and system components -- the result is in the our released code, in the \texttt{baselines/} folder. 

For retrieval backend: we replaced PaperQA's retrieval system with SearchR1's retrieval servers, using the same BM25 and E5 dense retrieval on the PubMed corpus that was used in all paper experiments. For LLM Integration: switched from proprietary APIs to local Qwen 2.5 models (3B/7B variants) served via vLLM, matching the exact models used in the SearchR1 training experiments. For answer format compatibility: we appended an instruction to the end of the PaperQA text prompt instructing the system to put the final single-entity answer into \texttt{<answer>} blocks, consistent with all the other content.  We created a standalone evaluation module, \texttt{qa\_em.py}, which matched the evaluation logic used in SearchR1 codebase (which is inside the \texttt{verl/} folder.

{
\onecolumn
\section{Training RLVR details}
\label{sec:appendix-training}
This section is single-column due to the large equation below. Continuing the description of the RL training algorithm from \cref{sec:training}, we leverage Group Relative Policy Optimization (GRPO) \cite{shao2024deepseekmath, guo2025deepseek}. At each iteration, we have the current policy $\pi_\theta$, which we now temporarily call the `old policy' $\pi_{old}$. For each question, $x$, GRPO computes multiple rollouts \( \{ y_1, y_2, \dots, y_G \} \) using $\pi_{old}$, and we can now consider some averaging of rewards ina group. The policy model is then optimized by maximizing:
{\scriptsize
\begin{align}\label{eq:grpo}
\mathcal{J}_{GRPO}(\theta) = \, & 
\mathbb{E}_{x \sim \mathcal{D}, \{ y_i \}_{i=1}^{G} \sim \pi_{\text{old}}( \cdot| x; \mathcal{R})}
\Bigg[
\frac{1}{G} \sum_{i=1}^{G} \frac{1}{\sum_{t=1}^{|y_i|}  I(y_{i,t})} \sum_{t=1: I(y_{i,t})=1}^{|y_i|} 
\min \Bigg( 
\frac{\pi_{\theta}(y_{i,t} | x, y_{i,<t}; \mathcal{R})}{\pi_{\text{old}}(y_{i,t} | x, y_{i,<t}; \mathcal{R})} \hat{A}_{i,t}, 
\nonumber \\[8pt] 
& \hspace{120pt} \text{clip} \Bigg( \frac{\pi_{\theta}(y_{i,t} | x, y_{i,<t}; \mathcal{R})}{\pi_{\text{old}}(y_{i,t} | x, y_{i,<t}; \mathcal{R})}, 1 - \epsilon, 1 + \epsilon \Bigg) \hat{A}_{i,t} 
\Bigg)
- \beta \mathbb{D}_{KL} \left[ \pi_{\theta} || \pi_{\text{ref}} \right]
\Bigg],
\end{align}
}
Here, $\mathcal{R}$ is the retriever (as before),  $\epsilon$ controls clipping range, and $\beta$ controls KL penalty. We compute `advantages' (rather than raw reward), $\hat{A}_{i,t}$ by normalizing rewards within each group of $G$ responses by using group mean as baseline and group standard deviation for scaling.

The full training scripts with all hyperparameters are available in the released code.

}

\end{document}